\documentclass[runningheads]{llncs}
\usepackage{graphicx}
\usepackage{amsmath,amssymb} 
\usepackage{color}

\usepackage{epsfig}
\usepackage{graphicx}
\usepackage{multirow}
\usepackage{xspace}
\usepackage{xfrac}
\usepackage{ifthen}
\usepackage{xcolor}

\usepackage{colortbl,dcolumn}

\usepackage{hhline}

\usepackage{pgfplots}
\usepackage{pgfplotstable}
\usepackage{pgf,tikz}

\newcommand{\extdata}[1]{\input{#1}}

\newcommand{\leg}[1]{\addlegendentry{#1}}

\usepackage[width=122mm,left=12mm,paperwidth=146mm,height=193mm,top=12mm,paperheight=217mm]{geometry}

\usepackage[pagebackref=true,breaklinks=true,letterpaper=true,colorlinks,bookmarks=false]{hyperref}

\begin{document}

\def\ECCV18SubNumber{2028}  

\title{Deep Shape Matching} 

\titlerunning{Deep Shape Matching}

\authorrunning{F. Radenovi\'c \and G. Tolias \and O. Chum}

\newcommand{\namespace}{\hspace{5mm}} \author{Filip Radenovi\'c \namespace Giorgos Tolias \namespace Ond\v{r}ej Chum}

\institute{Visual Recognition Group, FEE, CTU in Prague \\ \email{ \{filip.radenovic, giorgos.tolias, chum\}@cmp.felk.cvut.cz}}

\maketitle

\newcommand{\real}{\mathbb{R}}
\newcommand{\integer}{\mathbb{Z}}
\def\l2{\ensuremath{\ell_2}\xspace}
\def\linf{\ensuremath{\ell_\infty}\xspace}

\def\p{\ensuremath{p}\xspace}
\def\cI{\mathcal{I}}
\def\cS{\mathcal{S}}
\def\y{\mathbf{y}}
\def\x{\mathbf{x}}

\def\sssp{\hspace{1pt}}
\def\ssp{\hspace{3pt}}
\def\msp{\hspace{5pt}}
\def\bsp{\hspace{12pt}}

\def\etal{\emph{et al.}\xspace}
\def\ie{\emph{i.e.}\xspace}
\def\eg{\emph{e.g.}\xspace}
\def\etc{\emph{etc.}\xspace}

\newcommand{\alert}[1]{{\color{red}{#1}}}
\newcommand{\gio}[1]{{\color{purple}{#1}}}
\newcommand{\fr}[1]{{\color{blue}{#1}}}
\newcommand{\och}[1]{{\color{brown}{#1}}}
\newcommand{\comment}[1]{}

\renewcommand{\paragraph}[1]{\vspace{.4\baselineskip}\noindent{\bf #1}\xspace}
\newcommand{\paragraphitem}[1]{\vspace{.0\baselineskip}\noindent{\underline{\textit{#1}}}\xspace}

\newcommand{\xcaption}[2][1]{\caption{#2}\vspace{-#1\baselineskip}}

\newcommand{\eq}{eqn.\xspace}

\newcommand{\our}{$\boldsymbol{\star}$\xspace}

\definecolor{greenn}{rgb}{0.30,0.69,0.31}

\newcommand{\soaf}[1]{{\textbf{\color{red}{#1}}}}
\newcommand{\soas}[1]{{\textbf{\color{black}{#1}}}}
\newcommand{\soat}[1]{{\textbf{\color{blue}{#1}}}}

\newenvironment{itemizes}{%
\begin{list} {$\bullet$} {\setlength{\itemsep}{0pt}
                          \setlength{\leftmargin}{15pt}
                          \setlength{\topsep}{3pt} } }
{\end{list}\vspace{5pt}}

\newcommand\blfootnote[1]{%
  \begingroup
  \renewcommand\thefootnote{}\footnote{#1}%
  \addtocounter{footnote}{-1}%
  \endgroup
}

\begin{abstract}
We cast shape matching as metric learning with convolutional networks. We break the end-to-end process of image representation into two parts. Firstly, well established efficient methods are chosen to turn the  images into edge maps. Secondly, the network is trained with edge maps of landmark images, which are automatically obtained by a structure-from-motion pipeline. The learned representation is evaluated on a range of different tasks, providing improvements on challenging cases of domain generalization, generic sketch-based image retrieval or its fine-grained counterpart. In contrast to other methods that learn a different model per task, object category, or domain, we use the same network throughout all our experiments, achieving state-of-the-art results in multiple benchmarks.

\keywords{shape matching \and cross-modal recognition and retrieval}

\end{abstract}

\vspace{-10pt}
\section{Introduction}
\label{sec:intro}

\begin{figure}[b!]
\vspace{-5pt}
\input{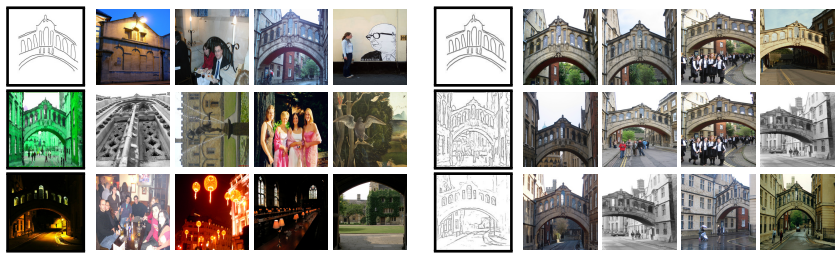}
\vspace{-20pt}
\caption{Three examples where {\bf shape} is the only relevant information: sketch, artwork, extreme illumination conditions. Top retrieved images from the Oxford Buildings dataset \cite{PCISZ07}: CNN with an RGB input~\cite{RTC16} (left), and our shape matching network (right). Query images are shown with black border.
\label{fig:bos}
}
\end{figure}

Deep neural networks have recently become very popular for computer-vision problems, mainly due to their good performance and generalization. These networks have been first used for image classification by Krizhevsky \etal~\cite{KSH12}, then their application spread to other related problems. 
A standard architecture of a classification network starts with convolutional layers followed by fully connected layers.
Convolutional neural networks (CNNs) became a popular choice of learning image embeddings, \eg in efficient image matching -- image retrieval. It has been observed that the convolutional part of the classification network captures well \emph{colours}, \emph{textures} and \emph{structures} within the receptive field.

In a number of problems, the colour and/or the texture is not available or misleading. Three examples are shown in Figure~\ref{fig:bos}. For sketches or outlines, there is no colour or texture available at all. For artwork, the colour and texture is present, but often can be unrealistic to stimulate certain impression rather than exactly capture the reality. Finally, in the case of extreme illumination changes, such as a day-time versus night images, the colours may be significantly distorted and the textures weakened.
On the other hand, the image discontinuities in colour or texture, as detected by modern edge detectors, and especially their shapes, carry the information about the content, independent of, or insensitive to, the illumination changes, artistic drawing and outlining. 

This work is targeting at shape matching, in particular the goal is to extract a descriptor that captures the shape depicted in the input. The shape descriptors are extracted from image edge maps by a CNN. Sketches, black and white line drawings or cartoons are simply considered as a special type of an edge map.

\begin{figure}[t]
\input{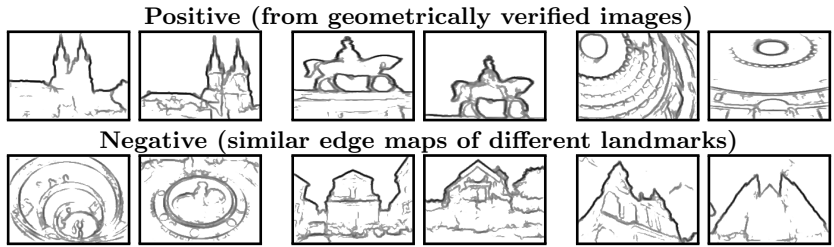}
\vspace{-15pt}
\caption{Edge maps extracted from matching and non-matching image pairs that serve as training data for our network.
\label{fig:intro}
\vspace{-15pt}
}
\end{figure}

The network is trained without any human supervision or image, sketch or shape annotation. Starting from a pre-trained classification network stripped off the fully connected layers, the CNN is fine-tuned using a simple contrastive loss function. 
Matching and non-matching training pairs are extracted from automatically generated 3D models of landmarks~\cite{RTC16}. Edge maps detected on these images provide training data for the network. Examples of positive and negative pairs of edge maps are shown in \mbox{Figure~\ref{fig:intro}}.

We show the importance of shape matching on two problems: 1) modality invariant representation, \ie classification for domain generalization, and 2) cross modality matching of sketches to images. 

In the domain generalization, some of the domains are available, but some are completely unseen during the training phase. We evaluate on domain generalization by performing object recognition.
We extract the learned descriptors and train a simple classifier on the seen domains, which is later used to classify images of the unseen domain.
We show, that for some combinations of seen-unseen domains, such as artwork and photograph, descriptors using colour and texture are useful. However, for some combinations, such as photograph and line drawing, the shape information is crucial. Combining both types of descriptors outperforms the state-of-the-art approach in all settings.

In the cross modality matching task, it is commonly assumed that annotated training data is available for both modalities. Even for this task, we apply the domain generalization approach, using the descriptors learned on edge maps of building images.
We evaluate the cross modality matching on sketch based image retrieval datasets.
Modern sketch-based image retrieval take the path of object recognition from human sketches. Rather than performing shape matching, the networks are trained to recognize simplified human drawings. Such an approach requires very large number of annotated images and drawn sketches for each category of interest.
Even though the proposed network is \emph{not trained} to recognize human-drawn object sketches, our experiments show that it performs well on standard benchmarks.

\vspace{-5pt}
\section{Related work}
\label{sec:related}
\vspace{-5pt}
Shape matching is shown useful in several computer vision tasks such as object recognition~\cite{BMJ02}, object detection~\cite{FFJS08}, 3D shape matching~\cite{WKL15,TL15} and cross-modal retrieval~\cite{CWZZ11,SWXZ13}.
In this section we review prior work related to sketch-based image retrieval, a particular flavor of cross-modal retrieval, where we apply the proposed representation. Finally, we discuss domain generalization approaches since our method is directly applicable on this problem handling it simply by learning shape matching. 

\paragraph{Sketch-based image retrieval}
has been, until recently, handled with hand-crafted descriptors~\cite{ERBHA12,RC13,Saaverda14,PM14,WZHM15,BC15,QSXZ+15,SBO15,XYHS+17,TC17}. 
Deep learning methods have been applied to the task of sketch-based retrieval~\cite{BYHR16,QSZL16,YLSX+16,SBHH16,SDM17,LSSLS17,BRPC16} much later than to the related task of image retrieval. We attribute the delay to the fact that the training data acquisition for sketch-based retrieval is much more tedious compared to image-based retrieval because it not only includes labeling the images, but also sketches must be drawn in large quantities.
Methods with no learning typically carry no assumptions on the depicted categories, while the learning based methods often include category recognition into training. The proposed method aims at generic sketch-based retrieval, not limited to a fixed set of categories; it is, actually, not even limited to objects.

\emph{Learning-free methods} have followed the same initial steps as in the traditional image search.
These include the construction of either global~\cite{CNM05,Saaverda14,QSXZ+15} or local~\cite{EHBA10,RDB11,RC13,CZL+13,WZHM15} image and/or sketch representations. Local representations are also using vector quantization to create a Bag-of-Words model~\cite{MYZR+13}.
Further cases are symmetry-aware and flip invariant descriptors~\cite{CZL+13}, descriptors that are based on local contours~\cite{RDB11} or line segments~\cite{WZHM15}, and 
 kernel descriptors~\cite{TC17}. Transformation invariance is often sacrificed for the sake of scalability~\cite{CWZZ11,SWXZ13}. In contrast, the method proposed in this paper is fully translation invariant, and scalable, because it reduces to a nearest-neighbor search in a descriptor space.

\emph{Learning-based methods} require annotated data in both domains, typically for a fixed set of object categories, making the methods~\cite{WKL15,BYHR16,QSZL16,YLSX+16,SBHH16,SDM17,LSSLS17,BRPC16,SYSXH17} to be category specific. 
End-to-end learning methods are applied to both category level~\cite{LSSLS17,BRPC16} and to fine-grained, \ie sub-category level retrieval~\cite{YLSX+16,SBHH16,SDM17,SYSXH17}, 
while sometimes a different model per category has to be learned~\cite{LHSG14,YLSX+16,SSZHR16,SYSXH17}.
A sequence of different learning and fine-tuning stages is applied~\cite{YLSX+16,SBHH16,SDM17,LSSLS17,BRPC16}, involving massive manual annotation effort. 
For example, the Sketchy dataset~\cite{SBHH16} required collectively 3,921 hours of sketching. 
On the contrary, our proposed fine-tuning does not require any manual annotation.

\paragraph{Domain generalization} is handled in a variety of ways, ranging from learning domain invariant features~\cite{GBZ+15,MBS13} to learning domain invariant classifiers~\cite{XLN+14,KZM+12} or both~\cite{LYS+17,BTS+16}. Several methods focus on one-way shift between two domains, such as sketch-based retrieval described earlier or learning on real photos and testing on art~\cite{CZ14,CZ16}. An interesting benchmark is released in the work of Li \etal~\cite{LYS+17}, where four domains of increasing visual abstraction are used, namely \emph{photos}, \emph{art}, \emph{cartoon}, and \emph{sketches} (PACS). Prior domain generalization methods~\cite{GBZ+15,MBS13,XLN+14} are shown effective on PACS, while simply training a CNN on all
the available (seen) domains is a very good baseline~\cite{LYS+17}. 
We tackle this problem from the representation point of view and focus on the underlined shapes.  
Our shape descriptor is extracted and the labels are used only to train a linear classifier.
In this fashion, we are able to train on a single domain and test on all the rest, while common domain generalization approaches require different domains present in the training set.
\vspace{-5pt}
\section{Method}
\label{sec:method}
\vspace{-5pt}
In this section we describe the proposed approach. The process of fine-tuning the CNN is described in Section~\ref{sec:methodT}, while the final representation and the way it is used for retrieval and classification is detailed in Section~\ref{sec:methodS}.

We break the end-to-end process of image description into two parts. In the first part, the images are turned into edge maps.
In particular, throughout all our experiments we used the edge detector of Doll{\'a}r and Zitnick~\cite{DZ13} due to its great trade-off between efficiency and accuracy, and the tendency not to consider textured regions as edges.
Our earlier experiments on sketch-based image retrieval with a CNN-based edge detector~\cite{Kok16} did not show any significant changes in the performance.
An image is represented as an edge map, which is a 2D array containing the edge strength in each image pixel. The edge strength is in the range of $[0, 1]$, where $0$ represents background. 
Sketches, in the case of sketch-to-image retrieval, are represented as a special case of an edge map, where the edge strength is either $0$ for the background or $1$ for a contour.

The second part is a fully convolutional network extracting a global image descriptor. 
The two part approach allows, in a simple manner, to unify all modalities at the level of edge maps. 
Jointly training these two parts, \eg in the case of a CNN-based edge detector~\cite{Kok16}, can deliver an image descriptor too. 
However, this descriptor may not be based on shapes. 
It is unlikely that such an optimization would end in a state where the representation between the two parts actually corresponds to edges. 
Enforcing this with additional training data in the form of edge maps and a loss on the output of the first part is exactly what we are avoiding and improving in this work.

\subsection{Training} 
\label{sec:methodT}

We use a network architecture previously proposed for image classification~\cite{SZ14}, in particular, we use all convolutional layers and the activations of the very last one, \ie, the network is stripped of the fully-connected layers. The CNN is initialized by the parameters learned on a large scale annotated image dataset, such as ImageNet~\cite{DSLLF09}. This is a fairly standard approach adopted in a number of problems, including image search~\cite{AGT+15,RTC16,GARL16}. 
The network is then fine-tuned with pairs of image edge maps.

\paragraph{The network.}
The image classification network expects an RGB input image, while the edge maps are only two dimensional. We sum the first convolution filters over RGB.
Unlike in RGB input, no mean pixel subtraction is performed to the input data.
To obtain a compact, shift invariant descriptor, a global max-pooling~\cite{RSAC14} layer is appended after the last convolutional layer.
This approach is also known as Maximum Activations of  Convolutions (MAC) vector~\cite{TSJ15}.
After the MAC layer, the vectors are \l2 normalized. 

\begin{figure*}[b]
\setlength{\fboxsep}{0pt} \setlength{\fboxrule}{1pt}
\begin{center}

\begin{tabular}{c@{\sssp}c}
\begin{tabular}{ccc}
\includegraphics[height=40pt, width=62pt]{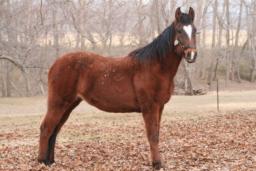} &
\fbox{\includegraphics[height=40pt, width=62pt]{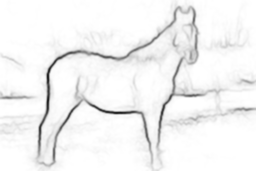}} &
\fbox{\includegraphics[height=40pt, width=62pt]{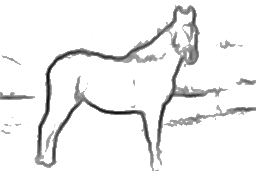}} 
\\
\includegraphics[height=40pt, width=62pt]{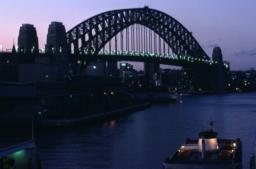} &
\fbox{\includegraphics[height=40pt, width=62pt]{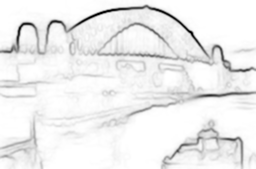}} &
\fbox{\includegraphics[height=40pt, width=62pt]{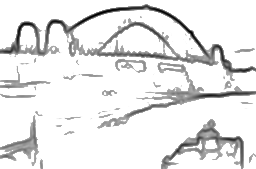}} 
\\
\end{tabular}
&
\hspace{0pt}
\raisebox{-50pt}{
\begin{tikzpicture}
\begin{axis}[%
    width=0.3\columnwidth,
    height=0.3\columnwidth,
    ylabel={$f(w)$},
    xlabel={$w$},
    tick label style  = {font=\small},
    ymin = -0.1,
    ymax = 1,
    xmin = 0,
    xmax = 1,
    xtick={0, 0.2, ...,1},        
    ytick={0, 0.2, ...,1},               
    grid=both,
    ylabel style={yshift=-3ex},      
    xlabel style={yshift=1ex},
    samples at={-0.1,-.09,...,1}
]
\addplot [blue, solid, line width=2.0, no marks] {x^0.4947 / (1 + exp(500*(0.0985-x)))}; 
\end{axis}
\end{tikzpicture}
}
\\
\end{tabular}
\end{center}
\vspace{-22pt}
\caption{Sample images, the output of the edge detector, the filtered edge map, and the edge-filtering function.
\label{fig:edgeproc}
}
\end{figure*}
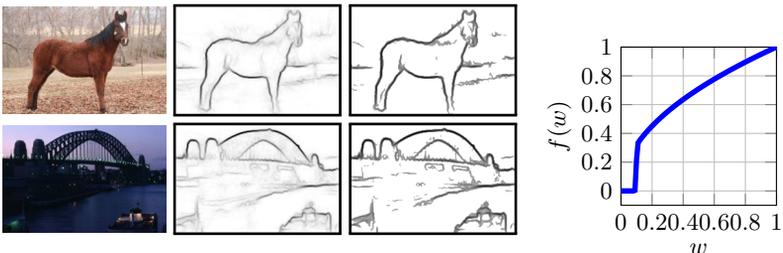

\paragraph{Edge filtering.}	
A typical output of edge detectors is a strength of an edge in every pixel. We introduce an edge filtering layer to address two frequent issues with edge responses. First, the background often contains close-to-zero responses, which typically introduce noise into the representation. This issue is commonly handled by thresholding the response function. Second, the strength of the edges provides ordering, \ie higher edge response implies that the edge is more likely to be present, however its value typically does not have practical interpretation. 
Prior to the first convolution layer, a continuous and differentiable function is pre-pended. This layer is trained together with the rest of the network to transform the edge detector output with soft thresholding by a sigmoid and power transformation.
Denote the edge strength by $w \in [0, 1]$. Edge filtering is performed as
\begin{equation}
f(w) = \frac{w^p}{1+e^{\beta(\tau-w)}}, \label{eqn:edgefilter}
\end{equation}
where $p$ controls the contrast between strong and weak edges, $\tau$ is the threshold parameter, and $\beta$ is the scale of the sigmoid choosing between hard thresholding and a softer alternative.
The final function (\ref{eqn:edgefilter}) with learned parameters is plotted in Figure~\ref{fig:edgeproc}~(right). The figure also visually demonstrates the effect of application of the filtering. The weak edges are removed on the background and the result appearance is closer to a rough sketch, while the uncertainty in edges is still preserved.

\paragraph{Fine tuning.}
The CNN is trained with Stochastic Gradient Descent in a Siamese fashion with contrastive loss~\cite{CHL05}.
The positive training pairs are edge maps of matching images (similarity of the edge maps is not considered), while the negative pairs are similar edge maps (according to the current state of the network) of non-matching images.

Given a pair of vectors $\x$ and $\y$, the loss is defined as their squared Euclidean distance $||\x-\y||^2$ for positive examples, and as $\max\{(m-||\x-\y||)^2, 0\}$ for negative examples. 
Hard-negative mining is performed several times per epoch which has been shown to be essential~\cite{RTC16,GARL16}.

\paragraph{Training data.}
The training images for fine tuning the network are collected in a fully automatic way.
In particular, we use the publicly available dataset used in Radenovic \etal~\cite{RTC16} and follow the same methodology, briefly reviewed in the following.
A large unordered image collection is passed through a 3D reconstruction system based on local features and Bag-of-Words retrieval~\cite{SRCF15,RSJ+16}.
The outcome consists of a set of 3D models which mostly depict outdoor landmarks and urban scenes.
For each landmark, a maximum of 30 six-tuples of images are being selected. The six-tuple consists of: one image as the training query, then one matching image to the training query, and five similar non-matching images. This gives arise to one positive and five negative pairs. 
The geometry of the 3D models, including camera positions, allows to mine matching images, \ie those that share adequate visual overlap. 
Negative-pair mining is facilitated by the 3D models, too: negative images are chosen only if they belong to a different model.

\paragraph{Data augmentation.}
A standard data-augmentation, \ie random horizontal flipping (mirroring) procedure is applied to introduce further variance in the training data and to avoid over-fitting.
The training query and the positive example are jointly mirrored with $50$\% probability.
Negative examples are sought after eventual flipping.
We propose an additional augmentation technique for the selected training queries.
Their edge map responses are thresholded with a random threshold uniformly chosen from $\left[0,0.2\right]$ and the result is binarized. Matching images (in positive examples) are left unchanged; negative images are selected after the transformation. This augmentation process is applied with a probability of $50$\%. It offers a level of shape abstraction and mimics the asymmetry of sketch-to-edge map matching. The randomized threshold can be also seen as an approximation of the stroke removal in~\cite{YLSX+16}.

\subsection{Representation, Search and Classification} \label{sec:methodS}
We use the trained network to extract image and sketch descriptors capturing the underlying shapes, which are then used to perform cross-modal image retrieval, in particular sketch-based, and object recognition via transfer learning, in particular domain generalization.

\paragraph{Representation.}
The input to the descriptor extraction process is always resized to a maximum dimensionality of $227 \times 227$ pixels.
A multi-scale representation is performed by processing at 5 fixed scales, \ie, re-scaling the original input by a factor of \sfrac{1}{2}, \sfrac{1}{$\sqrt{2}$}, 1, $\sqrt{2}$, 2, and, with the additional mirroring, 10 final instances are produced. 
\emph{Images} undergo edge detection and the resulting edge map~\cite{DZ13} is fed to the CNN\footnote{We perform zero padding by 30 pixels to avoid border effects.}.
\emph{Sketches} come in the form of strokes, thin line drawings, or brush drawings, depending on the input device or the dataset. To unify the sketch input, a simple morphological filter is applied to a binary sketch image. Specifically, a morphological thinning followed by dilation is performed. After the pre-processing, the sketch is treated as an edge map. 
As a consequence of the rescaling and mirroring, an image/sketch is mapped to 10 high dimensional vectors.
We refer to these \l2 normalized vectors as EdgeMAC descriptors.
They are subsequently sum-aggregated or indexed separately, depending on the evaluation benchmark, see Section~\ref{sec:experiments} for more details. 

\paragraph{Search.}
An image collection is indexed by simply extracting and storing the corresponding EdgeMAC descriptors for each image.
Search is performed by nearest-neighbors search of the query descriptor in the database.
This makes retrieval compatible with approximate methods~\cite{ML09,JDJ17} that can speed up search and offer memory savings. 

\paragraph{Classification.}
We extract EdgeMAC descriptors from labeled images and train a multi-class linear classifier~\cite{PAH+12} to perform object recognition. 
This is especially useful for transfer learning when the training domain is different from the target/testing one. In this case, no labeled images of the training domain are available during the training of our network and no labeled images of the target domain are available during classifier training. 
	
\subsection{Implementation details} 
\label{sec:training_implementation}
In this section we discuss implementation details. The training dataset used to train our network is presented. We train a single network, which is then used for different tasks. Training sets provided for specific tasks are not exploited.

\paragraph{Training data.}%
We use the same training set as in the work of Radenovic~\etal~\cite{RTC16}\footnote{Training data available at \href{http://cmp.felk.cvut.cz/cnnimageretrieval}{cmp.felk.cvut.cz/cnnimageretrieval}}
which comprises landmarks and urban scenes.
There are around 8k tuples.
Due to the overlap of landmarks contained in the training set and one of the test sets involved in our evaluation, we manually excluded these landmarks from our training data.
We end up with with 5,969 tuples for training and 1,696 for validation.
Hard negatives are re-mined 3 times per epoch~\cite{RTC16} from a pool of around 22k images.

\paragraph{Training implementation.}%
We use the MatConvNet toolbox~\cite{VL14} to implement the learning. 
We initialize the convolutional layers by VGG16~\cite{SZ14} (results in 512D EdgeMAC descriptor) trained on ImageNet and 
sum the filters of the first layer over the feature maps dimension to accommodate for the 2D edge map input instead of the 3D image.
The edge-filtering layer is initialized with values $p=0.5$, $\tau=0.1$ and $\beta$ is fixed and equal to 500 so that it always approximates hard thresholding. 
Additionally, the output of the egde-filtering layer is linearly scaled from $\left[0,1\right]$ to $\left[0,10\right]$.
Initial learning rate is $l_0=0.001$ with an exponential learning rate decay $l_0 \textrm{exp} (-0.1 j)$ over epoch $j$; momentum is 0.9; weight decay is 0.0005; contrastive loss margin is 0.7; and batch size is equal to 20 training tuples. 
All training images are resized so that the maximum extent is $200$ pixels, while keeping the original aspect ratio. 

\paragraph{Training time.}%
Training is performed for at most 20 epochs and the best network is chosen based on the performance on validation tuples. 
The whole training takes about 10 hours on a single GeForce GTX TITAN X (Maxwell) GPU with 12GB of memory.
\section{Experiments}
\label{sec:experiments}

We evaluate EdgeMAC descriptor on domain generalization and sketch-based image retrieval. 
We train a single network and apply it on both tasks proving the generic nature of the representation.

\subsection{Domain Generalization through Shape Matching}
We extract EdgeMAC descriptors from labeled images, sum-aggregate descriptors of rescaled and mirrored instances and \l2 normalize to produce one descriptor per image, and train a linear classifier~\cite{PAH+12} to perform object recognition.
We evaluate on domain generalization to validate the effectiveness of our representation on shape matching.

\paragraph{PACS dataset} was recently introduced by Li \etal~\cite{LYS+17}. 
It consists of images coming from 4 domains with varying level of abstraction, namely, \emph{art} (painting), \emph{cartoon}, \emph{photo}, and \emph{sketch}.
Images are labeled according to 7 categories, namely, dog, elephant, giraffe, guitar, horse, house, and person.
Each time, one domain is considered unseen, otherwise called target or test domain, while the image of the other 3 are used for training. 
Finally, multi-class accuracy is evaluated on the unseen domain.
In our work, we additionally perform classifier training using a single domain and then test on the rest. 
We find this scenario to be realistic, especially in the case of training on photos and testing on the rest. 
The domain of realistic photos is the richest in terms of annotated data, while others such as sketches and cartoons are very sparsely annotated.

\paragraph{Baselines.}
We are interested in translation invariant representations and consider the two following baselines. 
First, MAC~\cite{TSJ15} descriptors extracted using a network that is pre-trained on ImageNet.
Second, MAC descriptors extracted by a network that is fine-tuned for image retrieval in a siamese manner~\cite{RTC16}.
These two baselines have the same descriptor extraction complexity as ours.
They are extracted on RGB images, while ours on edge maps. Note, that we treat all domains as images with our approach and extract edge maps, \ie we do not perform any special treatment on sketches as in the case of sketch retrieval.

\begin{table}[t]
\caption{Multi-class accuracy on PACS dataset for 4 different descriptors. The combined descriptor (pre-trained + ours) is constructed via concatenation. A: Art, C: Cartoon, P: Photo, S: Sketch, 3: all 3 other domains.
\label{tab:pacs_score}
\vspace{-15pt}
}
\definecolor{greenish}{rgb}{0.3,0.7,0.3}
\definecolor{redish}{rgb}{0.7,0.3,0.3}
\definecolor{blueish}{rgb}{0.3,0.3,0.7}
\definecolor{someish}{rgb}{0.3,0.4,0.5}

\newcommand{\colg}[3]{%
	\newcommand*{\MinNumber}{#2}%
	\newcommand*{\MaxNumber}{#3}%
	\pgfmathsetmacro{\PercentColor}{100.0*(#1-\MinNumber+0.0*(\MaxNumber-\MinNumber))/(\MaxNumber-\MinNumber+0.0*(\MaxNumber-\MinNumber))}%
    \xdef\PercentColor{\PercentColor}%
    \cellcolor{greenish!\PercentColor}{#1}
}
\newcommand{\colr}[3]{%
  \newcommand*{\MinNumber}{#2}%
  \newcommand*{\MaxNumber}{#3}%
  \pgfmathsetmacro{\PercentColor}{100.0*(#1-\MinNumber+0.0*(\MaxNumber-\MinNumber))/(\MaxNumber-\MinNumber+0.0*(\MaxNumber-\MinNumber))}%
    \xdef\PercentColor{\PercentColor}%
    \cellcolor{redish!\PercentColor}{#1}
}
\newcommand{\colb}[3]{%
  \newcommand*{\MinNumber}{#2}%
  \newcommand*{\MaxNumber}{#3}%
  \pgfmathsetmacro{\PercentColor}{100.0*(#1-\MinNumber+0.0*(\MaxNumber-\MinNumber))/(\MaxNumber-\MinNumber+0.0*(\MaxNumber-\MinNumber))}%
    \xdef\PercentColor{\PercentColor}%
    \cellcolor{blueish!\PercentColor}{#1}
}
\newcommand{\cols}[3]{%
  \newcommand*{\MinNumber}{#2}%
  \newcommand*{\MaxNumber}{#3}%
  \pgfmathsetmacro{\PercentColor}{100.0*(#1-\MinNumber+0.0*(\MaxNumber-\MinNumber))/(\MaxNumber-\MinNumber+0.0*(\MaxNumber-\MinNumber))}%
    \xdef\PercentColor{\PercentColor}%
    \cellcolor{someish!\PercentColor}{#1}
}

\newcommand\cw{0.7cm}
\newcolumntype{C}[1]{>{\centering\let\newline\\\arraybackslash\hspace{-3pt}}m{#1}}

\begin{center}
\resizebox{\textwidth}{!}{%
\footnotesize
\begin{tabular}{lc@{\hspace{3pt}}c@{\hspace{3pt}}c@{\hspace{3pt}}c}
\begin{tabular}{l}
\\Test $\rightarrow$ \\Train A\\Train C\\Train P\\Train S\\Train 3\\Mean 3\\
\end{tabular}
&
\begin{tabular}{C{\cw}C{\cw}C{\cw}C{\cw}}
    \multicolumn{4}{c}{Pre-trained (RGB)}\\
    A & C & P & S \\
    N/A                &  \colr{59.2}{15}{70} &  \colb{95.0}{25}{110} &  \cols{33.1}{20}{90} \\
   \colg{71.7}{20}{80} &      N/A            &   \colb{86.8}{25}{110} &  \cols{37.0}{20}{90} \\ 
   \colg{72.5}{20}{80} &  \colr{33.3}{15}{70} &      N/A            &   \cols{24.8}{20}{90} \\ 
   \colg{31.9}{20}{80} &  \colr{49.5}{15}{70} &  \colb{42.5}{25}{110} &      N/A \\ \hline
      78.0 & 68.0 & 94.4 & 47.1 \\
    \multicolumn{4}{c}{71.9}\\      
\end{tabular}
&
\begin{tabular}{C{\cw}C{\cw}C{\cw}C{\cw}}
    \multicolumn{4}{c}{Siamese~\cite{RTC16} (RGB)}\\
    A & C & P & S \\
    N/A                &  \colr{59.5}{15}{70} &  \colb{86.3}{25}{110} &  \cols{42.9}{20}{90} \\
   \colg{61.0}{20}{80} &      N/A            &   \colb{77.0}{25}{110} &  \cols{51.6}{20}{90} \\ 
   \colg{66.0}{20}{80} &  \colr{38.0}{15}{70} &      N/A            &   \cols{31.9}{20}{90} \\ 
   \colg{38.7}{20}{80} &  \colr{49.3}{15}{70} &  \colb{44.4}{25}{110} &      N/A \\ \hline
    71.5 & 64.3 & 85.1 & 56.0 \\
    \multicolumn{4}{c}{69.2}\\      
\end{tabular}
&
\begin{tabular}{C{\cw}C{\cw}C{\cw}C{\cw}}
    \multicolumn{4}{c}{Ours (edge map)}\\
    A & C & P & S \\
    N/A                &  \colr{55.9}{15}{70} &  \colb{61.2}{25}{110} &  \cols{65.6}{20}{90} \\
   \colg{45.2}{20}{80} &      N/A            &   \colb{57.3}{25}{110} &  \cols{74.8}{20}{90} \\ 
   \colg{45.4}{20}{80} &  \colr{42.3}{15}{70} &      N/A            &   \cols{46.3}{20}{90} \\ 
   \colg{34.8}{20}{80} &  \colr{63.0}{15}{70} &  \colb{43.3}{25}{110} &      N/A \\ \hline
   53.8 & 67.9 & 64.5 & 74.7 \\
    \multicolumn{4}{c}{65.2}\\      
\end{tabular}
&
\begin{tabular}{C{\cw}C{\cw}C{\cw}C{\cw}}
    \multicolumn{4}{c}{Pre-trained\hspace{-1pt}+\hspace{-1pt}Ours}\\
    A & C & P & S \\
    N/A                &  \colr{61.6}{15}{70} &  \colb{94.9}{25}{110} &  \cols{38.4}{20}{90} \\
   \colg{69.3}{20}{70} &      N/A            &   \colb{85.0}{25}{110} &  \cols{55.3}{20}{90} \\ 
   \colg{73.3}{20}{70} &  \colr{34.0}{15}{70} &      N/A            &   \cols{27.61}{20}{90} \\ 
   \colg{33.7}{20}{70} &  \colr{59.3}{15}{70} &  \colb{43.4}{25}{110} &      N/A \\ \hline
    80.0 &  68.7 & 93.7 &  62.7 \\
   \multicolumn{4}{c}{76.2}\\      
\end{tabular}
\end{tabular}%
}
\end{center}

\vspace{-10pt}
\end{table}
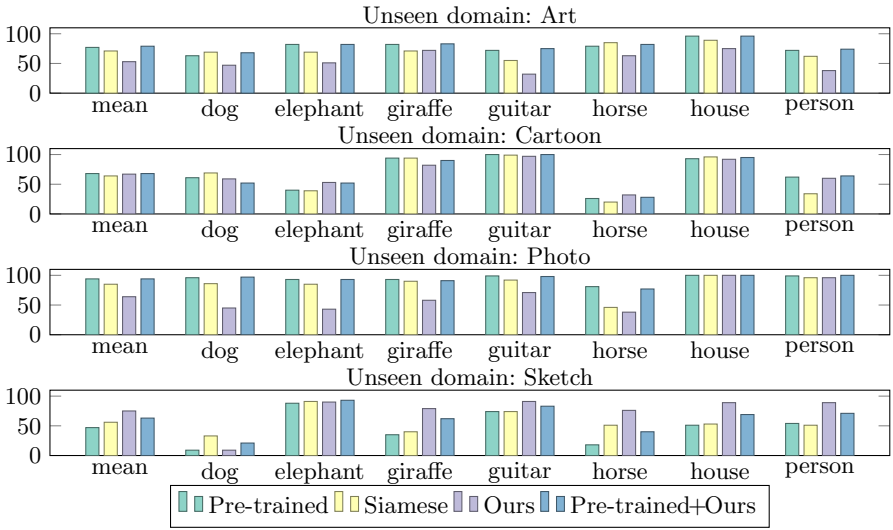
\begin{figure}[t]
\definecolor{C1}{RGB}{141, 211, 199}
\definecolor{C2}{RGB}{255, 255, 179}
\definecolor{C3}{RGB}{190, 186, 218}
\definecolor{C4}{RGB}{128, 177, 211}

\pgfplotsset{
  bar cycle list/.style={
    cycle list={%
      {C1!50!black,fill=C1,mark=none},%
      {C2!50!black,fill=C2,mark=none},%
      {C3!50!black,fill=C3,mark=none},%
      {C4!50!black,fill=C4,mark=none},%
    }
  },
}
%
\begin{tikzpicture}
\begin{axis}[
    ybar,
    title = {Unseen domain: Art},
    title style={yshift=-1.9ex},          
    enlarge y limits=upper,
    width=1.05\linewidth,
    height=0.2\linewidth,
    legend style={at={(0.5,-0.25)},anchor=north,legend columns=-1},
    symbolic x coords={mean, dog, elephant, giraffe, guitar, horse, house, person},
    xtick=data,
    /pgf/bar width=5pt,
    xtick style={draw=none},
    ymin = 0, ymax = 100,
    ytick={0,50,...,100},
    x tick label style={inner sep=-2pt},    
    ]
\addplot coordinates {(mean, 77) (dog, 63) (elephant, 82) (giraffe, 82) (guitar, 72) (horse, 79) (house, 96) (person, 72)};
\addplot coordinates {(mean, 71) (dog, 69) (elephant, 69) (giraffe, 71) (guitar, 55) (horse, 85) (house, 89) (person, 62)};
\addplot coordinates {(mean, 53) (dog, 47) (elephant, 51) (giraffe, 72) (guitar, 32) (horse, 63) (house, 75) (person, 38)};
\addplot coordinates {(mean, 79) (dog, 68) (elephant, 82) (giraffe, 83) (guitar, 75) (horse, 82) (house, 96) (person, 74)};
\end{axis}
\end{tikzpicture}
%
%

%
\begin{tikzpicture}
\begin{axis}[
    ybar,
    title = {Unseen domain: Cartoon},
    title style={yshift=-1.9ex},          
    enlarge y limits=upper,
    width=1.05\linewidth,
    height=0.2\linewidth,
    legend style={at={(0.5,-0.25)},anchor=north,legend columns=-1},
    symbolic x coords={mean, dog, elephant, giraffe, guitar, horse, house, person},
    xtick=data,
    /pgf/bar width=5pt,
    xtick style={draw=none},
    ymin = 0, ymax = 100,
    ytick={0,50,...,100},
    x tick label style={inner sep=-2pt},        
    ]
\addplot coordinates {(mean, 68) (dog, 61) (elephant, 40) (giraffe, 94) (guitar, 100) (horse, 26) (house, 93) (person, 62)};
\addplot coordinates {(mean, 64) (dog, 69) (elephant, 39) (giraffe, 94) (guitar, 99) (horse, 20) (house, 96) (person, 34)};
\addplot coordinates {(mean, 67) (dog, 59) (elephant, 53) (giraffe, 82) (guitar, 97) (horse, 32) (house, 92) (person, 60)};
\addplot coordinates {(mean, 68) (dog, 52) (elephant, 52) (giraffe, 90) (guitar, 100) (horse, 28) (house, 95) (person, 64)};
\end{axis}
\end{tikzpicture}
%
%

%
\begin{tikzpicture}
\begin{axis}[
    ybar,
    title = {Unseen domain: Photo},
    title style={yshift=-1.9ex},          
    enlarge y limits=upper,
    width=1.05\linewidth,
    height=0.2\linewidth,
    legend style={at={(0.5,-0.25)},anchor=north,legend columns=-1},
    symbolic x coords={mean, dog, elephant, giraffe, guitar, horse, house, person},
    xtick=data,
    /pgf/bar width=5pt,
    xtick style={draw=none},
    ymin = 0, ymax = 100,
    ytick={0,50,...,100},
    x tick label style={inner sep=-2pt},        
    ]
\addplot coordinates {(mean, 94) (dog, 96) (elephant, 93) (giraffe, 93) (guitar, 99) (horse, 81) (house, 100) (person, 99)};
\addplot coordinates {(mean, 85) (dog, 86) (elephant, 85) (giraffe, 90) (guitar, 92) (horse, 46) (house, 100) (person, 96)};
\addplot coordinates {(mean, 64) (dog, 45) (elephant, 43) (giraffe, 58) (guitar, 71) (horse, 38) (house, 100) (person, 96)};
\addplot coordinates {(mean, 94) (dog, 97) (elephant, 93) (giraffe, 91) (guitar, 98) (horse, 77) (house, 100) (person, 100)};
\end{axis}
\end{tikzpicture}
%
%

%
\begin{tikzpicture}
\begin{axis}[
    ybar,
    title = {Unseen domain: Sketch},
    title style={yshift=-1.9ex},          
    enlarge y limits=upper,
    width=1.05\linewidth,
    height=0.2\linewidth,
    legend style={at={(0.5,-0.45)},anchor=north,legend columns=-1},
    symbolic x coords={mean, dog, elephant, giraffe, guitar, horse, house, person},
    xtick=data,
    /pgf/bar width=5pt,
    xtick style={draw=none},
    ymin = 0, ymax = 100,
    ytick={0,50,...,100},
    x tick label style={inner sep=-2pt},        
    ]
\addplot coordinates {(mean, 47) (dog, 09) (elephant, 88) (giraffe, 35) (guitar, 74) (horse, 18) (house, 51) (person, 54)};
\addplot coordinates {(mean, 56) (dog, 33) (elephant, 91) (giraffe, 40) (guitar, 74) (horse, 51) (house, 53) (person, 51)};
\addplot coordinates {(mean, 75) (dog, 09) (elephant, 90) (giraffe, 79) (guitar, 91) (horse, 76) (house, 89) (person, 89)};
\addplot coordinates {(mean, 63) (dog, 21) (elephant, 93) (giraffe, 62) (guitar, 83) (horse, 40) (house, 69) (person, 71)};
\legend{Pre-trained, Siamese, Ours, Pre-trained\hspace{-1pt}+\hspace{-1pt}Ours}
\end{axis}
\end{tikzpicture}
\vspace{-7pt}
\caption{Classification accuracy on PACS dataset with different descriptors.
Testing is performed on 1 unseen domain each time, while training is performed on the other 3.\label{fig:pacs_category}
\vspace{-10pt}}
\end{figure}

\begin{figure}[t]
\vspace{-5pt}
\begin{center}
\begin{tabular}{c@{\hspace{40pt}}c}
\hspace{50pt}
\begin{tikzpicture}
\scriptsize
\begin{axis}[
    title = {Pre-trained (RGB)},
    width=0.35\linewidth,
    height=0.35\linewidth,
    ymin = -75, ymax = 65,
    xmin = -75, xmax = 65,    
    ticks=none,
    title style = {yshift = -8pt},
    ]
\addplot[only marks, mark=*,mark size=.4,color=red, each nth point={2}] table[x index=0,y index=1]{./figs/data/pretrain_domain1_tsne.txt};
\addplot[only marks, mark=*,mark size=.4,color=green, each nth point={2}] table[x index=0,y index=1]{./figs/data/pretrain_domain2_tsne.txt};
\addplot[only marks, mark=*,mark size=.4,color=blue, each nth point={2}] table[x index=0,y index=1]{./figs/data/pretrain_domain3_tsne.txt};
\addplot[only marks, mark=*,mark size=.4,color=magenta, each nth point={2}] table[x index=0,y index=1]{./figs/data/pretrain_domain4_tsne.txt};
\end{axis}
\end{tikzpicture}
&
\begin{tikzpicture}
\scriptsize
\begin{axis}[
    title = {Ours (edge map)},
    width=0.35\linewidth,
    height=0.35\linewidth,
    legend style={opacity = 0.7, font=\tiny, row sep = -2pt, mark options={scale=3.5}, xshift = 50pt},
    legend cell align={right},
    legend pos=south east,
    ymin = -55, ymax = 55,
    xmin = -55, xmax = 55,    
    ticks=none,
    title style = {yshift = -8pt},
    ]
\addplot[only marks, mark=*,mark size=.4,color=red, each nth point={2}] table[x index=0,y index=1]{./figs/data/edgemac_domain1_tsne.txt};
\addplot[only marks, mark=*,mark size=.4,color=green, each nth point={2}] table[x index=0,y index=1]{./figs/data/edgemac_domain2_tsne.txt};
\addplot[only marks, mark=*,mark size=.4,color=blue, each nth point={2}] table[x index=0,y index=1]{./figs/data/edgemac_domain3_tsne.txt};
\addplot[only marks, mark=*,mark size=.4,color=magenta, each nth point={2}] table[x index=0,y index=1]{./figs/data/edgemac_domain4_tsne.txt};
\legend{Art, Cartoon, Photo, Sketch};
\end{axis}
\end{tikzpicture}
\\
\end{tabular}
\end{center}
\vspace{-20pt}
\caption{Visualization of PACS images with t-SNE (more overlap is better).\label{fig:pacs_tsne}
\vspace{-32pt}}
\end{figure}
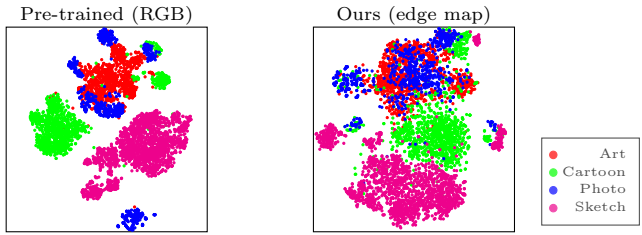

\paragraph{Performance comparison.}
We evaluate our descriptor, the two baselines, and the concatenated version of ours and the descriptor of the pre-trained baseline network, and report results in Table~\ref{tab:pacs_score}.
Our representation significantly improves sketch recognition while training on a single or all seen domains.
Similar improvements are observed for cartoon recognition when training on photos or sketches, while when training on artwork the color information appears to be beneficial. 
We consider the case of training only on photos and testing on other domains to be the most interesting and realistic one.
In this scenario, we provide improvements, compared to the baselines, for sketch recognition (15\% and 22\%) and cartoon recognition (4\% and  9\%). 
Finally, the combined descriptor reveals the complementarity of the representations in several cases, such as artwork and cartoon recognition while training on all seen domains, or training on single domain when artwork is involved, \eg. train on P (or A) and test on A (or C).
The best reported score on PACS is 69.2~\cite{LYS+17} by fine-tuning AlexNet on PACS.
The achieved score by our descriptor with fine-tuned VGG (PACS not used during network training) is 76.2, which is significantly higher. The same experiment with AlexNet achieves 70.9.
Performance is reported per category in Figure~\ref{fig:pacs_category}. Our descriptor achieves significant improvements on most categories for sketch recognition, while the combined is a safe choice in several cases.
Interestingly, our experiments reveal that the siamese baseline slightly improves shape matching, despite being trained on RGB images.

\paragraph{Visualization with t-SNE.}
We use t-SNE~\cite{MG08} to reduce the dimensionality of descriptors to 2 and visualize the result for the pre-trained baseline and our descriptor in Figure~\ref{fig:pacs_tsne}. The different modalities are brought closer with our descriptor. Observe how separated is the sketch modality with the pre-trained network that receives an RGB image for input.

\subsection{Sketch-based Image Retrieval}
\label{sec:experiments_sbir}
We extract EdgeMAC descriptors to index an image collection, treat sketch queries as described in Section~\ref{sec:methodS} and perform sketch-based image retrieval via simple nearest neighbor search.

\paragraph{Test datasets and evaluation protocols.} The method is evaluated on two standard sketch-based image retrieval benchmarks.

\emph{Flickr15k}~\cite{RC13} consists of 15k database images and 330 sketch queries that are related to 33 categories. 
Categories include particular object instances (Brussels Cathedral, Colosseum, Arc de Triomphe, \etc), generic objects (airplane, bicycle, bird, \etc), and shapes (circle shape, star shape, heart, balloon, \etc).
The performance is measured via mean Average Precision (mAP)~\cite{PCISZ07}. 
We sum-aggregate descriptors of rescaled and mirrored instances and \l2 normalize to produce one descriptor per image.
Search is performed by a cosine similarity nearest-neighbor search.

\emph{Shoes/Chairs/Handbags}~\cite{YLSX+16,SYSXH17} datasets contain images from one category only, \ie shoe/chair/handbag category respectively. 
It consists of pairs of a photo and a corresponding hand-drawn detailed sketch of this photo.
There are 419, 297, and 568 sketch--photo pairs of shoes, chairs, and handbags, respectively.
Out of these, 304, 200, and 400 pairs are selected for training, and 115, 97, and 168  for testing shoes, chairs, and handbags, respectively.
The performance is measured via the matching accuracy at the top~K retrieved images, denoted by acc.@K.
The underlying task is quite different compared to Flickr15k. The photograph used to generate the sketch is to be retrieved, while all other images are considered false positives.
The search protocol used in~\cite{YLSX+16} is as follows: Descriptors are extracted from 5 image crops (corners and center) and their horizontally mirrored counterparts. This holds for database images and the sketch query. During search, these 10 descriptors are compared one-to-one and their similarity is averaged. For fair comparison, we adopt this protocol and do not use a single descriptor per image/sketch for this benchmark. However, instead of image crops, we extract descriptors at 5 image scales and their horizontally mirrored counterparts, as these are defined in Section~\ref{sec:methodS}.

\begin{table}[t!]
\caption{Performance evaluation of the different components of our method on Flickr15k dataset. Network: off-the-shelf~(O), fine-tuned~(F).
\label{tab:exp_components}
\vspace{-10pt}
}
\newcolumntype{L}[1]{>{\raggedright\let\newline\\\arraybackslash\hspace{0pt}}m{#1}}
\newcolumntype{C}[1]{>{\centering\let\newline\\\arraybackslash\hspace{0pt}}m{#1}}
\newcolumntype{R}[1]{>{\raggedleft\let\newline\\\arraybackslash\hspace{0pt}}m{#1}}
\newcommand\cw{1cm}
\newcommand\bsw{0.65cm}
\def\arraystretch{1.0}
\newcommand{\bs}{{\tiny{$\blacksquare$}}} 
\footnotesize
\begin{center}
\setlength{\tabcolsep}{4pt}
\setlength\extrarowheight{-1pt}
\begin{tabular}{|@{~~}L{4cm}|C{\bsw}|C{\bsw}|C{\bsw}|C{\bsw}|C{\bsw}|C{\bsw}|C{\bsw}|C{\bsw}|}
    \hline
    \multirow{2}{*}{Component} & \multicolumn{8}{c|}{Network} \\
    \cline{2-9}
    & \raisebox{-2pt}{O} & \raisebox{-2pt}{O} & \raisebox{-2pt}{F} & \raisebox{-2pt}{F} & \raisebox{-2pt}{F} & \raisebox{-2pt}{F} & \raisebox{-2pt}{F} & \raisebox{-2pt}{F} \\
    \hline
    Train/Test: Edge filtering & & \bs & \bs & \bs & \bs & \bs & \bs & \bs \\
    Train: Query binarization & & & & \bs & \bs & \bs & \bs & \bs \\
    Test: Mirroring & & & & & \bs & & \bs & \bs \\
    Test: Multi-scale & & & & & & \bs & \bs & \bs \\
    Test: Diffusion & & & & & & & & \bs \\
    \hline\hline
    \raisebox{-2pt}{mAP} & \raisebox{-2pt}{25.9} & \raisebox{-2pt}{27.9} & \raisebox{-2pt}{38.4} & \raisebox{-2pt}{41.9} & \raisebox{-2pt}{43.4} & \raisebox{-2pt}{45.6} & \raisebox{-2pt}{46.1} & \raisebox{-2pt}{68.9} \\
    \hline
\end{tabular}
\end{center}
\vspace{-25pt}
\end{table}

\paragraph{Impact of different components.} Table~\ref{tab:exp_components} shows the impact of different components on the final performance of the proposed method as measured on Flickr15k dataset. Direct application of the off-the-shelf CNN on edge maps already outperforms most prior hand-crafted methods (see Table~\ref{tab:exp_flickr15k}). Adding the edge-filtering layer to the off-the-shelf network improves the precision. The initial parameters for filtering are used.
Fine-tuning brings significant jump to $38.4$ mAP, which is already the state-of-the-art on this dataset. 
Random training-query binarization and multi-scale with mirroring representation further improve the mAP score to $46.1$.
Finally, we boost the recall of our sketch-based retrieval by global diffusion, as recently proposed by Iscen~\etal~\cite{ITA+16}.
We construct the neighborhood graph by combining kNN-graphs built on two different similarities~\cite{BSX+16,ZYCYM12}: edge map similarity with EdgeMAC and image similarity with CroW descriptors~\cite{KMO15}. This increases the performance to $68.9$.

\begin{figure*}[t]
\vspace{-6pt}
\centering
\input{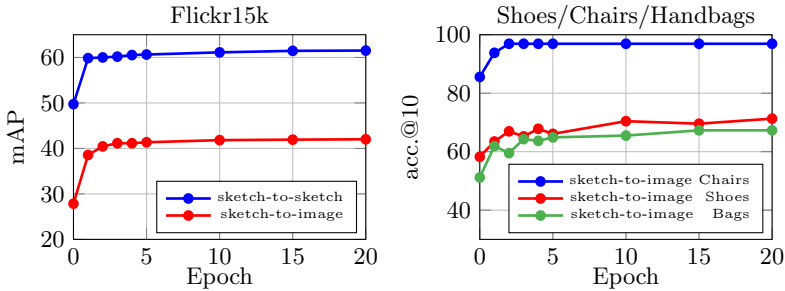}
\begin{tabular}{cc}
\begin{tikzpicture}
\begin{axis}[%
	ylabel near ticks, yticklabel pos=left,
	xlabel near ticks,
	font=\footnotesize,
	width=0.45\linewidth,
	height=0.35\linewidth,
	xlabel={Epoch},
	ylabel={mAP},
	xlabel style  = {yshift = 5pt},
	title={Flickr15k},
	legend pos=south east,
    legend style={cells={anchor=east}, font =\tiny, fill opacity=1, row sep=-2.5pt},	
    ymax = 65,
    ymin = 20,
    xmin = 0,
    xmax = 20,
    grid=both,
    xtick={0,5,...,20},    
    ytick={20, 30, ..., 100},
    title style = {yshift = -5pt}
]

	\addplot[color=blue, solid, mark=*, mark size=1.5, line width=1.0] table[x=epoch, y expr={100*\thisrow{sk2sk}}] \flickr; \leg{sketch-to-sketch};
	\addplot[color=red, solid, mark=*, mark size=1.5, line width=1.0] table[x=epoch, y expr={100*\thisrow{sk2im}}] \flickr; \leg{sketch-to-image};

\end{axis}
\end{tikzpicture}
& 
\begin{tikzpicture}
\begin{axis}[%
	ylabel near ticks, yticklabel pos=left,
	xlabel near ticks,
	font=\footnotesize,
	width=0.45\linewidth,
	height=0.35\linewidth,
	xlabel={Epoch},
	ylabel={acc.@10},
	xlabel style  = {yshift = 5pt},
	title={Shoes/Chairs/Handbags},
	legend pos=south east,
    legend style={cells={anchor=east}, font =\tiny, fill opacity=1, row sep=-2.5pt},	
    ymax = 100,
    ymin = 30,
    xmin = 0,
    xmax = 20,
    grid=both,
    xtick={0,5,...,20},    
    ytick={0,20,...,100},
    title style = {yshift = -7pt}
]

	\addplot[color=blue, solid, mark=*, mark size=1.5, line width=1.0] table[x=epoch, y expr={\thisrow{chairs@10}}] \shoeschairs; \leg{sketch-to-image Chairs};
	\addplot[color=red, solid, mark=*, mark size=1.5, line width=1.0] table[x=epoch, y expr={\thisrow{shoes@10}}]  \shoeschairs; \leg{sketch-to-image {~}Shoes};
    \addplot[color=greenn, solid, mark=*, mark size=1.5, line width=1.0] table[x=epoch, y expr={\thisrow{handbags@10}}]  \shoeschairs; \leg{sketch-to-image {~~}Bags};

\end{axis}
\end{tikzpicture}
\end{tabular}
\vspace{-13pt}
\caption{Performance evaluation of the fine-tuned network over training epochs for the single-scale representation.
 All shown datasets and their evaluation protocols are described in Section~\ref{sec:experiments_sbir}.
\label{fig:exp_ep_vs_res}
\vspace{-5pt}
}
\end{figure*}

\paragraph{Performance evolution during learning.} We report the performance of the fine-tuned network at different stages (epochs) of training. The same network is evaluated for all datasets as we train a single network for all tasks. 
The performance is shown in Figure~\ref{fig:exp_ep_vs_res} for both benchmarks.
On all datasets, the fine-tuning significantly improves the performance already from the first few epochs.

As a sanity check, we also perform a non-standard sketch-to-sketch evaluation. On the Flickr15k dataset, each of the 330 sketches is used to query the other 329 sketches (the query sketch is removed from the evaluation), which attempts to retrieve sketches of the same category. 
The evolution of the performance shows similar behavior as the sketch-to-image search, \ie, the learning on edge maps improves the performance on sketch-to-sketch retrieval, see Figure~\ref{fig:exp_ep_vs_res}.

\paragraph{Comparison with the state of the art.} We extensively compare our method to the state-of-the-art performing methods on both benchmarks.
Whenever code and trained models are publicly available, we additionally evaluate them on test sets they were not originally applied on.
In cases that the provided code is used for evaluation on Flickr15k we center and align the sketches appropriately in order to achieve high scores, while our method is translation invariant so there is no such need.
First we give a short overview of the best performing and most relevant methods to ours.
Finally, a comparison via quantitative results is given.

\emph{Shoes/Chairs/Handbags networks}~\cite{YLSX+16,SYSXH17} are trained from scratch based on the Sketch-a-Net architecture~\cite{YYSXH15}.
This is achieved by the following steps~\cite{YLSX+16}\footnote{Networks/code available at \href{https://github.com/seuliufeng/DeepSBIR}{github.com/seuliufeng/DeepSBIR}}:
(i)~Training with classification loss for 1k categories from ImageNet-1K data with edge maps input.
(ii)~Training with classification loss for 250 categories of TU-Berlin~\cite{EHA12} sketch data.
(iii)~Training a triplet network with shared weights and ranking loss on TU-Berlin sketches and ImageNet images.
(iv)~Finally, training separate networks for fine-grain instance-level ranking using the Shoes/Chairs/ /Handbags training datasets.
This approach is later improved~\cite{SYSXH17} by adding an attention module with a coarse-fine fusion (CFF) into the architecture, and by extending the triplet loss with a higher order learnable energy function (HOLEF).
Such a training involves various datasets, with annotation at different levels, and a variety of task-engineered loss functions.
Note that the two models available online achieve higher performance than the ones reported in~\cite{YLSX+16}, due to parameter retuning.
We compare our results to their best performing models.

\emph{Sketchy network}~\cite{SBHH16} consists of two asymmetric sketch and image branches, both initialized with GoogLeNet.
The training involves the following steps\footnote{Network/code available at \href{https://github.com/janesjanes/sketchy}{github.com/janesjanes/sketchy}}:
(i)~Training for classification on TU-Berlin sketch dataset.
(ii)~Separate training of the sketch branch with classification loss on 125 categories of Sketchy dataset and training of the image branch with classification loss on the same categories with additional 1000 Flickr photos per category.
(iii)~Training both branches in a triplet network with ranking loss on the Sketchy sketch--photo pairs. 
The last part involves approximately 100k positive and a billion negative pairs.

\emph{Quadruplet network}~\cite{SDM17} tackles the problem in a similar way as Sketchy network, however, they use ResNet-18~\cite{HZRS16} architecture with shared weights for both sketch and image branches.
The training involves the following steps:
(i)~Training with classification loss on Sketchy dataset. 
(ii)~Training a network with triplet loss on Sketchy dataset, while mining three different types of triplets.

\emph{Triplet no-share network}~\cite{BRPC16} consists of asymmetric sketch and image branches initialized by Sketch-a-Net and AlexNet~\cite{KSH12}, respectively.
The training involves:
(i)~Separate training of the sketch branch with classification loss on TU-Berlin and training of the image branch with classification loss on ImageNet.
(ii)~Training a triplet network with ranking loss on TU-Berlin sketches augmented with 25k corresponding photos harvested from the Internet.
(iii)~Training a triplet network with ranking loss on the Sketchy dataset.

\paragraph{Performance comparison.} We compare our network with other methods on both benchmarks. Methods that have not reported scores on a particular datasets are evaluated by ourselves while using the publicly available networks.

Results on the Flickr15k dataset are presented in Table~\ref{tab:exp_flickr15k}, where our method significantly outperforms both hand-crafted descriptors and CNN-based that are learned on a variety of training data. This holds for both plain search with the descriptors, and for methods using re-ranking techniques, such as query expansion~\cite{CPSIZ07} and diffusion~\cite{ITA+16}.

\begin{table}[t]
\caption{Performance comparison via mean Average Precision~(mAP) with the state-of-the-art sketch-based image retrieval on the Flickr15k dataset. Best result is highlighted in \protect\soaf{red}, second best in \protect\soas{bold}. Query expansion methods are shown below the horizontal line and are highlighted separately. Our evaluation of the methods that do not originally report results on Flickr15 is marked with $^\dagger$.
\label{tab:exp_flickr15k}
\vspace{-5pt}
}
\newcolumntype{L}[1]{>{\raggedright\let\newline\\\arraybackslash\hspace{0pt}}m{#1}}
\newcolumntype{C}[1]{>{\centering\let\newline\\\arraybackslash\hspace{0pt}}m{#1}}
\newcolumntype{R}[1]{>{\raggedleft\let\newline\\\arraybackslash\hspace{0pt}}m{#1}}
\newcommand\cw{1cm}
\def\arraystretch{1.0}
\begin{center}

\begin{tabular}{cc}

\scriptsize
\setlength{\tabcolsep}{0mm}
\setlength\extrarowheight{-1pt}
\begin{tabular}{|@{~~}L{3cm}|R{0.6cm}@{~}|C{0.7cm}|}
    \multicolumn{3}{c}{\textbf{Hand-crafted methods}} \\ [2pt]
    \hline
    Method & Dim & mAP \\
    \hline \hline
    GF-HOG~\cite{RC13} & n/a & 12.2\\
    S-HELO~\cite{Saaverda14} & 1296 & 12.4\\
    HLR+S+C+R~\cite{WZHM15} & n/a & 17.1\\
    GF-HOG extended~\cite{BC15} & n/a & 18.2\\
    PerceptualEdge~\cite{QSXZ+15} & 3780 & 18.4\\
    LKS~\cite{SBO15} & 1350 & 24.5\\
    AFM~\cite{TC17} & 243 & 30.4\\ \hline
    AFM+QE~\cite{TC17} & 755 & \soas{57.9} \\
    \hline
\end{tabular}

&

\scriptsize
\setlength\extrarowheight{0pt}
\setlength{\tabcolsep}{0mm}
\begin{tabular}{|@{~~}L{5cm}|R{0.6cm}@{~}|C{0.7cm}|}
    \multicolumn{3}{c}{\textbf{CNN-based methods}} \\ [2pt]
    \hline
    Method & Dim & mAP \\
    \hline \hline
    Sketch-a-Net+EdgeBox~\cite{BYHR16} & 5120 & 27.0 \\
    Shoes network~\cite{YLSX+16}$^\dagger$  & 256 & 29.9 \\
    Chairs network~\cite{YLSX+16}$^\dagger$ & 256 & 29.8 \\
    Sketchy network~\cite{SBHH16}$^\dagger$ & 1024 & 34.0 \\
    Quadruplet network~\cite{SDM17} & 1024 & 32.2 \\
    Triplet no-share network~\cite{BRPC16} & 128 & \soas{36.2} \\
    \our EdgeMAC & 512 & \soaf{46.1} \\ \hline
    Sketch-a-Net+EdgeBox+GraphQE~\cite{BYHR16} & n/a & 32.3 \\
    \our EdgeMAC+Diffusion & n/a & \soaf{68.9} \\    
    \hline
\end{tabular}

\end{tabular}

\end{center}
\vspace{-15pt}
\end{table}

\begin{table}
\caption{Performance comparison via accuracy at rank K (acc.@K) with the state-of-the-art sketch-based image retrieval on the Shoes/Chairs test datasets. Best result is highlighted in \protect\soaf{red}, second best in \protect\soas{bold}.
Note that \cite{YLSX+16} and \cite{SYSXH17} train a separate network per object category.
$^\dagger$We evaluate the publicly available networks, because the performance is higher than the one originally reported in~\cite{YLSX+16}.
\label{tab:exp_chairsshoes}
\vspace{-10pt}
}
\newcolumntype{L}[1]{>{\raggedright\let\newline\\\arraybackslash\hspace{0pt}}m{#1}}
\newcolumntype{C}[1]{>{\centering\let\newline\\\arraybackslash\hspace{0pt}}m{#1}}
\newcolumntype{R}[1]{>{\raggedleft\let\newline\\\arraybackslash\hspace{0pt}}m{#1}}
\newcommand\cw{1cm}
\newcommand\ecw{1mm}
\def\arraystretch{0.95}
\begin{center}
\scriptsize
\setlength{\tabcolsep}{0mm}
\setlength\extrarowheight{0pt}
\begin{tabular}{|@{~}L{4.8cm}|R{0.7cm}@{~}|C{\ecw}|C{\cw}|C{\cw}|C{\ecw}|C{\cw}|C{\cw}|C{\ecw}|C{\cw}|C{\cw}|}
    \cline{1-2} \cline{4-5} \cline{7-8} \cline{10-11}
    \multirow{2}{*}{Method} & \multirow{2}{*}{Dim} & & \multicolumn{2}{c|}{Shoes} & & \multicolumn{2}{c|}{Chairs} & & \multicolumn{2}{c|}{Handbags} \\
    \cline{4-5} \cline{7-8} \cline{10-11}
    & & & acc.@1 & acc.@10 & & acc.@1 & acc.@10 & & acc.@1 & acc.@10 \\
	\hhline{==~==~==~==}
    BoW-HOG ~~~~+ rankSVM~\cite{YLSX+16} & 500 & & 17.4 & 67.8 & & 28.9 & 67.0 & & 2.4 & 10.7 \\
    Dense-HOG ~~~+ rankSVM~\cite{YLSX+16} & 200K & & 24.4 & 65.2 & & 52.6 & 93.8 & & 15.5 & 40.5 \\
    Sketch-a-Net ~~+ rankSVM~\cite{YLSX+16} & 512 & & 20.0 & 62.6 & & 47.4 & 82.5 & & 9.5 & 44.1 \\
    CCA-3V-HOG + PCA~\cite{XYHS+17} & n/a & & 15.8 & 63.2 & & 53.2 & 90.3 & & -- & -- \\
    Shoes ~~~~~net~\cite{YLSX+16}$^\dagger$ & 256 & & 52.2 & \soas{92.2} & & 65.0 & 92.8 & & 23.2 & 59.5 \\
    Chairs ~~~~net~\cite{YLSX+16}$^\dagger$ & 256 & & 30.4 & 75.7 & & 72.2 & \soaf{99.0} & & 26.2 & 58.3 \\
    Handbags net~\cite{SYSXH17} & 256 & & -- & -- & & -- & -- & & 39.9 & 82.1 \\
    Shoes ~~~~~net + CFF + HOLEF~\cite{SYSXH17} & 512 & & \soaf{61.7} & \soaf{94.8} & & -- & -- & & -- & -- \\
    Chairs ~~~~net + CFF + HOLEF~\cite{SYSXH17} & 512 & & -- & -- & & \soas{81.4} & 95.9 & & -- & -- \\
    Handbags net + CFF + HOLEF~\cite{SYSXH17} & 512 & & -- & -- & & -- & -- & & \soas{49.4} & \soas{82.7} \\
    \our EdgeMAC & 512 & & 40.0 & 76.5 & & \soaf{85.6} & 95.9 & & 35.1 & 70.8 \\
    \our EdgeMAC + whitening & 512 & & \soas{54.8} & \soas{92.2} & & \soaf{85.6} & \soas{97.9} & & \soaf{51.2} & \soaf{85.7} \\
    \cline{1-2} \cline{4-5} \cline{7-8} \cline{10-11}
\end{tabular}%
\end{center}
\vspace{-20pt}
\end{table}

Results on the fine-grained Shoes/Chairs/Handbags benchmark are shown in Table~\ref{tab:exp_chairsshoes}.
In this experiment, we also report the performance after applying descriptor whitening which is learned in a supervised way~\cite{RTC16} by using the descriptors of the training images of this benchmark. A single whitening transformation is learned for all three datasets. Such a process takes only a few seconds once descriptors are given. It is orders of magnitude faster than using the training set to perform network fine-tuning.
We achieve the top performance in 2 out of 3 categories and the second best in the other one. 
The approach of~\cite{YLSX+16} and~\cite{SYSXH17} train a separate network per category (3 in total), which is clearly not scalable to many objects. In contrast our approach uses a single generic network. 
An additional drawback is revealed when we evaluate the publicly available Shoes and Chairs networks on the category they were not trained on.
We observe a significant drop in performance, see Table~\ref{tab:exp_chairsshoes}.

\paragraph{The number of parameters.} Our reported results use the VGG16 network stripped off the fully connected layers (FC), leaving $\sim$15M parameters. 
The number of parameters of Sketch-A-Net~\cite{YYSXH15} is $\sim$8.5M parameters, while when used for SBIR in two different branches (Shoes, Chairs, Handbags~\cite{YLSX+16}) there is $\sim$17M parameters. 
Triplet no-share network~\cite{BRPC16} uses two branches (Sketch-a-Net with additional FC layer and AlexNet~\cite{KSH12}) leading to $\sim$115M, and Sketchy~\cite{SBHH16} uses 2$\times$ GoogLeNet leading to $\sim$26M parameters. Our network has the smallest number of parameters from the competing methods.

\vspace{-5pt}
\section{Conclusions}
\label{sec:conclusions}
\vspace{-5pt}

We have introduced an approach to learn shape matching by training a CNN with edge maps of matching images.
The training stage does not require any manual annotation, achieved by following the footsteps of image retrieval~\cite{RTC16}, where image pairs are  automatically mined from large scale 3D reconstruction\blfootnote{{\bf Acknowledgements}: This work was supported by the MSMT LL1303 ERC-CZ grant and the OP VVV funded project
 CZ.02.1.01/0.0/0.0/16\_019/0000765 ``RCI''.}. 

The generic applicability of the representation is proven by validating on a variety of cases. 
It achieves state-of-the-art results on standard benchmarks for sketch-based image retrieval, while 
we have further demonstrated the applicability beyond sketch-based image retrieval.
Promising results were achieved for queries with different modality (artwork) and significant change of illumination (day-night retrieval).
The descriptor is shown beneficial for object recognition via transfer learning, especially to classify images of unseen domains, such as cartoons and sketches, where the amount of annotated data is limited. 
Remarkably, the same network is applied in all the different tasks. 
Training data, trained models, and code are publicly available at \href{http://cmp.felk.cvut.cz/cnnimageretrieval}{cmp.felk.cvut.cz/cnnimageretrieval}.

{\small
\bibliographystyle{splncs}
\bibliography{egbib}
}

\end{document}